\documentclass[sigconf]{acmart}
\usepackage{booktabs} 
\setcopyright{rightsretained}
\usepackage{url}
\usepackage{latexsym}
\usepackage{amsfonts}
\usepackage{graphicx}
\usepackage{amsmath}
\usepackage{tabu}
\usepackage{algorithm}  
\usepackage{algorithmic}  
\usepackage{booktabs}
\usepackage{amssymb}
\usepackage{booktabs}
\usepackage{tabularx}
\usepackage{multirow}
\usepackage{bbm}
\usepackage{algorithm}  
\usepackage{algorithmic}  
\acmDOI{10.475/123_4}

\acmISBN{123-4567-24-567/08/06}
\acmYear{2018}
\copyrightyear{2018}
\acmArticle{4}
\acmPrice{15.00}

\begin{document}
\title{Dialogue Act Recognition via CRF-Attentive Structured Network}


\author{Zheqian Chen$^*$, Rongqin Yang$^*$, Zhou Zhao$^\dag$, Deng Cai$^*$, Xiaofei He$^*$}
\affiliation{
	\institution{	$^*$State Key Lab of CAD$\&$CG, College of Computer Science, Zhejiang University, Hangzhou, China\\
		$^\dag$College of Computer Science, Zhejiang University, Hangzhou, China\\
	}
$\{$zheqianchen, rongqin\_yrq,dengcai, xiaofeihe$\}$@gmail.com,    zhaozhou@zju.edu.cn\\
}
\begin{abstract}
Dialogue Act Recognition (DAR) is a challenging problem in dialogue interpretation, which aims to attach semantic labels to utterances and characterize the speaker's intention. Currently, many existing approaches formulate the DAR problem ranging from multi-classification to structured prediction, which suffer from handcrafted feature extensions and attentive contextual structural dependencies. In this paper, we consider the problem of DAR from the viewpoint of extending richer Conditional Random Field (CRF) structural dependencies without abandoning end-to-end training. We incorporate hierarchical semantic inference with memory mechanism on the utterance modeling. We then extend structured attention network to the linear-chain conditional random field layer which takes into account both contextual utterances and corresponding dialogue acts. The extensive experiments on two major benchmark datasets Switchboard Dialogue Act (SWDA) and Meeting Recorder Dialogue Act (MRDA) datasets show that our method achieves better performance than other state-of-the-art solutions to the problem. It is a remarkable fact that our method is nearly close to the human annotator's performance on SWDA within 2\% gap.
\end{abstract}

\maketitle

\begin{figure*}[t]
	\centering
	\includegraphics[width=16cm]{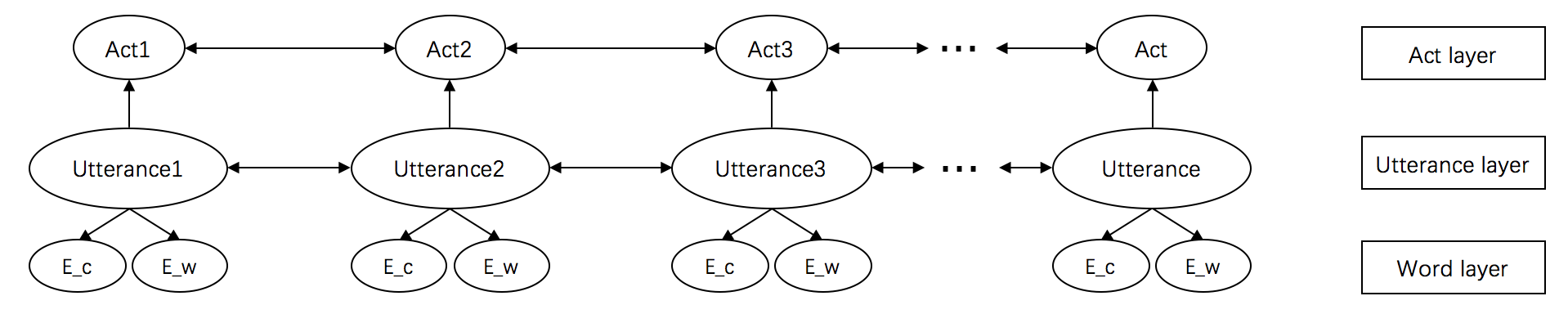}
	\caption{An illustration of the hierarchical conversation structure. The input of the model is a conversation which consist of $n$ utterances $u_1,u_2,...,u_n$ with corresponding dialogue act labels $a_1,a_2,...,a_n$. Each utterance is composed of diverse length of words in the character level $E_c$ and the word level $E_w$. Notice that utterances are not exist independent, utterances have contextual relations with each other.}
\end{figure*}

\section{Introduction}
Dialogue Act Recognition (DAR) is an essential problem in modeling and detecting discourse structure. The goal of DAR is to attach semantic labels to each utterance in a conversation and recognize the speaker's intention, which can be regarded as a sequence labeling task. Many applications have benefited from the use of automatic dialogue act recognition such as dialogue systems, machine translation, automatic speech recognition, topic identification and talking avatars~\cite{Khanpour2016Dialogue}~\cite{kral2012dialogue}~\cite{higashinaka2014towards}. One of the primary applications of DAR is to support task-oriented discourse agent system. Knowing the past utterances of DA can help ease the prediction of the current DA state, thus help to narrow the range of utterance generation topics for the current turn. For instance, the "Greeting" and "Farewell" acts are often followed with another same type utterances, the "Answer" act often responds to the former "Question" type utterance. Thus if we can correctly recognize the current dialogue act, we can easily predict the following utterance act and generate a corresponding response. Table 1 shows a snippet of the kind of discourse structure in which we are interested.

\begin{table}
	\begin{tabular}{ccl}
		\toprule
		Speaker    &Utterance       &DA Label\\
		\midrule
		A &Hi, long time no see.& Greeting\\
		B &Hi, how are you? & Greeting\\
		A &What are you doing these days? &Question \\
		B &I'm busying writing my paper. &Answer\\
		A &I heard that the deadline is coming. &Statement\\
		B &Yeah. &Backchannel \\
		A &You need to make a push.&Opinion \\
		B &Sure, that's why I am so busy now.&Agreement\\
		A &I can't bother you for too long, goodbye. &Farewell\\
		B &See you later. &Farewell\\
		\bottomrule
	\end{tabular}
\caption{ A snippet of a conversation sample. Each utterance has related dialogue act label.}
\end{table}

The essential problem of DAR lies on predicting the utterance's act by referring to contextual utterances with act labels. Most of existing models adopt handcrafted features and formulate the DAR as a multi-classification problem. However, these methods which adopt feature engineering process and multi-classification algorithms reveal deadly weakness from two aspects: First, they are labor intensive and can not scale up well across different datasets. Furthermore, they abandon the useful correlation information among contextual utterances. Typical multi-classification algorithms like SVM, Naive Bayes~\cite{grau2004dialogue}~\cite{Ang2005AutomaticDA}~\cite{stolcke2006dialogue} can not account for the contextual dependencies and classify the DA label in isolation. It is evident that during a conversation, the speaker's intent is influenced by the former utterance such as the previous "Greeting" and "Farewell" examples. To tackle these two problems, some works have turn to structured prediction algorithm along with deep learning tactics such as DRLM-Conditional~\cite{Ji2016ALV}, LSTM-Softmax~\cite{Khanpour2016Dialogue} and RCNN~\cite{Kalchbrenner2013Recurrent}. However, most of them failed to utilize the empirical effectiveness of attention in the graphical structured network and relies completely on the hidden layers of the network, which may cause the structural bias. A further limitation is that although these works claim they have considered the contextual correlations, in fact they view the whole conversation as a flat sequence and neglect the dual dependencies in the utterance level and act level~\cite{blunsom2013recurrent}~\cite{huangbidirectional}~\cite{maend}. Until now, the achieved performances in DAR field are still far behind human annotator's accuracy.

In this paper, we present the problem of DAR from the viewpoint of extending richer CRF-attentive structural dependencies along with neural network without abandoning end-to-end training. For simplicity, we call the framework as \textbf{CRF-ASN} (CRF-\textbf{A}ttentive \textbf{S}tructured \textbf{N}etwork). Specifically, we propose the hierarchical semantic inference integrated with memory mechanism on the utterance modeling. The memory mechanism is adopted in order to enable the model to look beyond localized features and have access to the entire sequence. The hierarchical semantic modeling learns different levels of granularity including word level, utterance level and conversation level. We then develop internal structured attention network on the linear-chain conditional random field (CRF) to specify structural dependencies in a soft manner. This approach generalizes the soft-selection attention on the structural CRF dependencies and takes into account the contextual influence on the nearing utterances. It is notably that the whole process is differentiable thus can be trained in an end-to-end manner. 

The main contributions of this paper are as follows:
\begin{itemize}
\item Unlike the previous studies, we study dialogue act recognition from the viewpoint of extending rich CRF-attentive structural dependencies. The proposed CRF structural attention on the DAR problem provides an alternative approach to encode the internal utterance inference with dialogue acts.

\item We propose the hierarchical deep recurrent neural network with memory enhanced mechanism to fully model the utterance semantic representations. The proposed framework can be trained end-to-end from scratch and can be easily extended across different dialogue tasks. 

\item We conduct extensive experiments on two popular datasets SWDA and MRDA to show that our method outperform several state-of-the-art solutions on the problem. It is worth noting that our method has achieved nearly close to the human annotator's performance on SWDA within 2\% gap, which is very convincing.
\end{itemize}

The rest of this paper is organized as follows. In section 2, we introduce the problem of dialogue act recognition from the viewpoint of introducing CRF-structured attention, and propose the CRF-attentive structural network with hierarchical semantic inference and memory mechanism. A variety of experimental results are presented in Section 3. We have a comprehensive analysis on the experiment results and conduct the ablations to prove the availability of our model. We then provide a brief review of the related work about dialogue act recognition problem in Section 4. Finally, we provide some concluding remarks in Section 5.  

\section{CRF-attentive Structured Network  }
In this section, we study the problem of dialogue act recognition from the viewpoint of extending rich CRF-attentive structural dependencies. We first present the hierarchical semantic inference with memory mechanism from three levels: word level, utterance level and conversation level. We then develop graphical structured attention to the linear chain conditional random field to fully utilize the contextual dependencies.

\begin{figure*}[t]
	\centering
	\includegraphics[width=17cm]{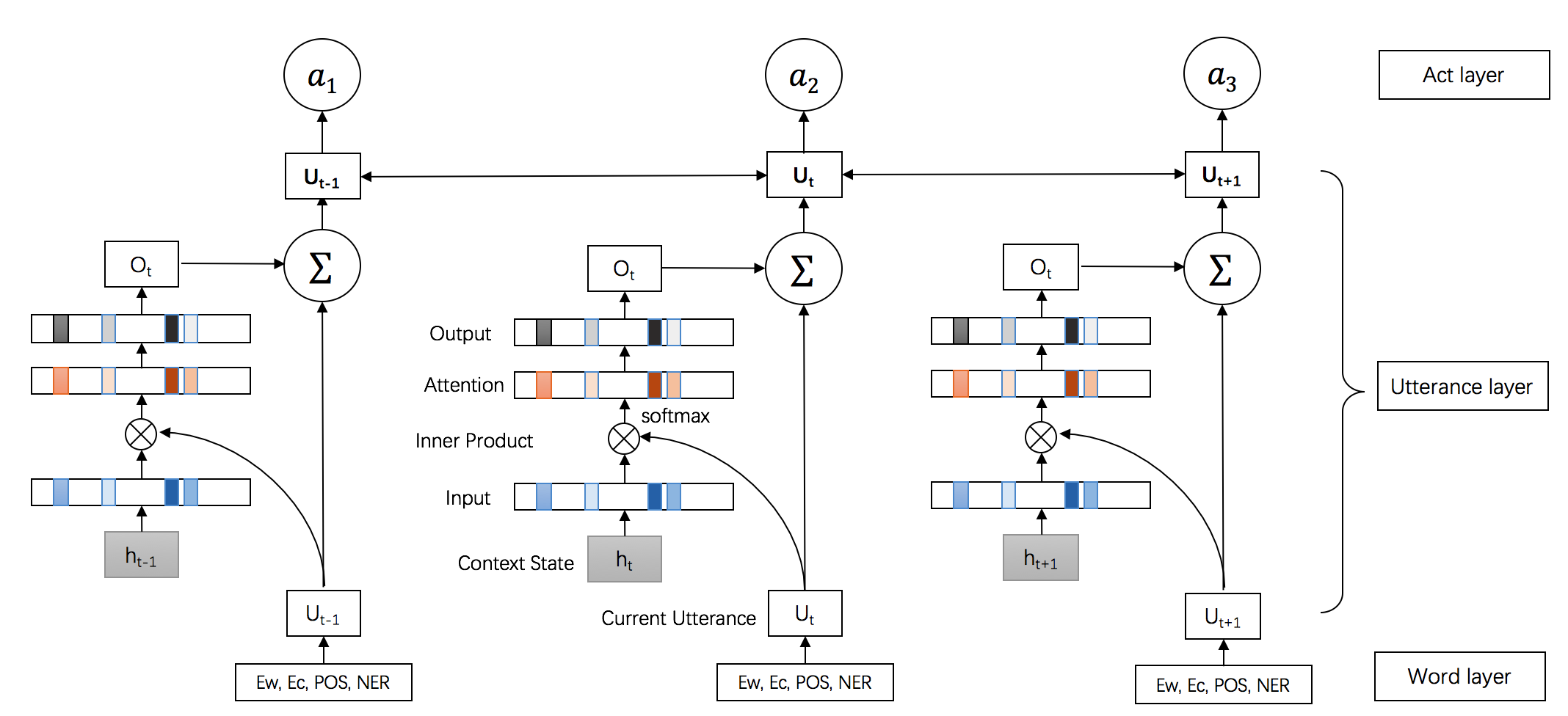}
	\caption{The Overview of learning memory enhanced hierarchical conversation representation architecture. The momory hop is set to 1. First concatenate the rich word embedding and obtain the original utterance representation $u_t$ from the basic BiGRU. The hidden state $h_t$ represents the contextual encoding which cares the former and the latter utterance dependencies. After summarizing hierarchical memory enhanced output $o_t$ and the original utterance $u_t$, we get the final representation $u_t$ denoted in a bold form.}
\end{figure*}

\subsection{The problem}
Before presenting the problem, we first introduce some basic mathematical notions and terminologies for dialogue act recognition. Formally, we assume the input is in the form of sequence pairs: $D=(C_1,C_2,...,C_{N})$ with $Y=(Y_1, Y_2,...,Y_M)$. $C_n$ is the input of the $n$-th conversation in dataset $D$ and $Y_m$ is the $m$-th targeted dialogue act type. Each conversation $C_i$ is composed of a sequence of utterances which denoted as $C_i=(u_1,u_2,...,u_j)$ with aligned act types $(y_1,y_2,...,y_j)$. We have each dialogue act type assigned to utterance $u_j \rightarrow y_j$ and each associated $y_j \in Y$ denoted the possible dialogue act belongs to $Y$ act types. Again each utterance consists of a sequence of diverse words $u_i=(w_1,w_2,...,w_t)$.

Most of the previous models do not leverage the implicit and intrinsic dependencies among dialogue act and utterances. They just consider a conversation as a flat structure with an extremely long chain of words. However, such a construction suffers vanishing gradient problem as the extremely long words become impractical in the neural network back-propagation training process. To alleviate this problem, we consider the conversation to be a hierarchical structure composed of three level encoders: first encode each word in a fine grained manner, and the second encoder operates at the utterance level, the last encoder encode each utterance in the conversation level. Each encoder is based on the previous one thus can make sure the output of the previous one can capture the dependencies across the conversation. Here we take an example to illustrate the sequence structure in Figure 1. Apart from hierarchical neural encoders, we also integrate external memory to allow the model to have unrestricted access to the whole sequence rather than localized features as in RNNs.

Naturally the dialogue act recognition problem can be regarded as a sequence labeling task which can be assigned dialogue act through multi-classification method or the structured prediction algorithms. In our formulation, we adopt the linear chain conditional random field (CRF) along with hierarchical attentive encoders for the structured prediction. Instead of labeling each utterance in isolation, structured prediction models such as HMM, CRF can better capture the contextual dependencies among utterances. In our model, we define the structured attention model as being an extended attention model which provides an alternative approach to incorporate the machinery of structural inference directly into our neural network. 

\subsection{Hierarchical Semantic Network}
Due to the hierarchical nature of conversations, our proposed model is constructed at multiple levels of granularity, e.g. word level, utterance level and conversation level. The representation of a conversation can be composed by each utterance $u_j$, and each $u_j$ can be obtained by combining the representations of constituent words $w_t$. Taking inspiration from Memory Networks and incorporate so-called memory hops, we adopt the memory enhanced contextual representations in order to have unrestricted access to the whole sequence rather than localized features as former recurrent neural network. Here we include the memory enhanced hierarchical representation in Figure 2 to depict the conversation level representation.

As illustrated in Figure 2, the hierarchical semantic network can be divided into two parts: (1) fine grained embedding layer (2) memory enhanced contextual representation layer. The second part can be further broken down into three main components: (a) the input memory $m_{1:t}$ which takes in the output from the  word embedding layer (b) the contextual attention which takes the consideration of the former utterance and the latter one. (c) the output memory $c_{1:t}$ which is obtained from the input memory connected with the attention mechanism. The weights are determined by measuring the similarity between the input memory and the current utterance input. 

\noindent \textbf{Fine Grained Embedding: } For a given conversation, each utterance $u_j$ is encoded by a fine grained embedding layer. We first try to utilize the rich lexical factors and linguistic properties to enhance the word representation. For each word token $w_t$ in each utterance, we initialized the word embedding using pretrained embeddings such as Word2vec or Glove. Furthermore, in order to tackle the out-of-vocabulary (OOV) problem, we adopt the character-level word embedding via CNN to combine with pretrained word level embeddings. We also extend the lexical factors via POS tag and NER tag to enhance the utterance understanding. The obtained four factors are concatenated to form a rich lexical representation as:
\begin{align*}
      e_k=f_{embed}(w_t,c_t,pos,ner) \tag{1}
\end{align*}

Since we consider the bidirectional GRU to encode the representation of each utterance, we concatenate the outputs from the forward and backward GRU hidden representations at the time step. For each utterance $u_j$ which consists a sequence of words $w_1,w_2,...,w_j$, the original semantic representation is as follows:
\begin{align*}
	u_j=\left\lbrace \overrightarrow{{f_{GRU}}}(h_{k-1},e_k),\overleftarrow{{f_{GRU}}}(h_{k-1},e_k)  \right\rbrace \tag{2}
\end{align*}   

Here we utilize $f_{embed}$ and $f_{GRU}$ to represent the word level embedding function and utterance level encoder in our hierarchical model. After obtained the original semantic representations on each utterance, we later apply the memory enhanced contextual layer to further explore the correlations between utterances. 

\noindent \textbf{Memory Enhanced Contextual Representation: } Every utterance in a conversation is encoded with $u_j=\Phi(e_k)$, where $\Phi\left (\cdot \right)$ is the encoding function via Bi-GRU to map the input words into a vector $u_j \in \mathbb{R}^d$. The original sequence utterances are denoted as $\left\lbrace u_1,u_2,...,u_j \right\rbrace $. While this original semantic representation can be the input component in the context of memory network. In order to tackle the drawback of insensitivity to temporal information between memory cells, we adopt the approach in injecting temporal signal into the memory using a contextual recurrent encoding:
\begin{align*}
	\overleftarrow{h_j}&=\overleftarrow{GRU}(u_j,\overleftarrow{h_{j-1}}) \\
	\overrightarrow{h_j}&=\overrightarrow{GRU}(u_j,\overrightarrow{h_{j-1}}) \\
	h_j&=tanh(\overleftarrow{W_m}\overleftarrow{h_j}+\overrightarrow{W_m}\overrightarrow{h_j}+b_m) \tag{3}
\end{align*} 
where $\overleftarrow{W_m}$, $\overrightarrow{W_m}$, $b_m$ are learnable parameters.

It is a remarkable fact that the new sequence $h_j$ can be seen as the contextual integrated representations which take consider of the former utterances and the latter ones. The injected temporal signal can further explore the contextual influence on the current input utterance. We thus can make use of this obtained $h_j$ to represent another $u_j$ which cares more about the context influence.

For the current input utterance $u_j$, in memory networks, the input is required to be in the same space as the input memory. Here we adopt the popular attention mechanism in the memory by measuring the relevance between current input utterance $u_j$ and the contextual new representation $h_j$. The relevance is measured with a softmax function:
\begin{align*}
	p_{j,i}=softmax(u_j^Th_i) \tag{4}
\end{align*}

Once the attention weights have been computed, the output memory can be used to generate the final output of the memory layer in the form of a weighted sum over the attention and the input utterance:
\begin{align*}
	o_t=\sum_i p_{j,i}u_j \tag{5}
\end{align*}

The output allows the model to have unrestricted access to elements in previous steps as opposed to a single hidden state $u_j$ in recurrent neural networks. Thereby we can effectively detect the long range dependencies among utterances in a conversation.

To further extend the complex reasoning over multiple supporting facts from memory, we adopt a stacking operation which stacks hops between the original utterance semantic representation $u_j$ and the k-th output hop $o_t$ to be the input to the $(k+1)$th hop:
\begin{align*}
	u_t^{k+1}=o_t^k+u_t^k \tag{6}
\end{align*} 
where $u_t^{k+1}$ encodes not only information at the current step ($u_j^k$), but also relevant knowledge from the contextual memory ($o_t^k$). Note that in the scope of this work, we limit the number of hops to 1 to ease the computational cost.

\subsection{Structured CRF-Attention Network}
Traditional attention networks have proven to be an effective approach for embedding categorical inference within a deep neural network. However, In DAR problem, we need to further explore the 
structural dependencies among utterances and dialogue acts. As we see, utterances in a conversation are not exist independently. The latter utterance may be the responding answer to the former question, or that the chunk of utterances are in the same act type. Here we consider generalizing selection to types of chunks selecting attention, and propose the structured attention to model richer dependencies by incorporating structural distributions within networks. Such a structured attention can be interpreted as using soft-selection that considers all possible structures over the utterance input. 

In our paper, we formulate the DAR as a sequence labeling problem. It is a natural choice to assign a label to each element in the sequence via linear chain CRF, which enable us to model dependencies among labels. Here we do not directly apply the original linear chain CRF to the learned utterance. Although the dependencies among utterances have been captured by the former hierarchical semantic networks, we still need to further explore the dialogue act dependencies in the label level. For dialogue act sequence labeling problem, greedily predicting the dialogue act at each time-step might not optimal the solution. Instead, it is better to look into the correlations in both utterance level and the dialogue act level in order to jointly decode the best chain of dialogue acts.

Formally, let $u=\left[u_1,u_2,...,u_n \right] $ represent a sequence of utterance inputs, let $y=\left[y_1,y_2,...,y_n\right]$ be the corresponding dialogue act sequence. Variable $z$ are discrete latent act variables $\left[ z_1,z_2,...,z_n\right] $ with sample space $ \left\lbrace 1,...,n\right\rbrace $ that encodes the desired selection among these inputs. The aim of the structured attention is to produce a sequence aware $conversation$ $c$ based on the utterances $u$ and the dialogue act sequence $y$. We assume the attentive distribution $z=p(z|u_i,y)$, where we condition $p$ on the input utterances $u$ and the dialogue act sequence $y$.  Here we assume the utterances in the conversation as an undirected graph structure with $n$ vertices. The CRF is parameterized with clique potentials $\theta_c(z_c) \in \mathbb{R}$, indicating the subset of $z$ give by clique $c$. Under this definition, the attention probability is defined as $p(z|u,q;\theta)=softmax(\sum_c\theta_c(z_c))$. For symmetry, we use the softmax in a general sense, i.e. $softmax(g(z))=\frac{1}{z}exp(g(z))$, where $z=\sum_{z'}exp(g(z'))$ is the implied recognition function. Here $\theta$ comes from the former memory enhanced deep model over utterances $u$ and corresponding dialogue acts $y$. 

The $conversation$ $c$ over the utterances and dialogue acts is defined as expectation:
\begin{align*}
c=\mathbb{E}_{z\sim p(z|u,y)}[f(u,z)]=\sum_c \mathbb{E}_{z\sim p(z_c|u,y)}[f_c(u,z_c)] \tag{7}
\end{align*}
where we assume the annotation function $f$ factors into $f(u,z)=\sum_c f_c(u,z_c)$. The annotation function is defined to simply return the selected hidden state. The $conversation$ $c$ can be interpreted as an dialogue act aware attentive conversation as taking the expectation of the annotation function with respect to a latent variable $z\sim p$, where $p$ is parameterized to be function of utterances $u$ and dialogue acts $y$.
 
The expectation is a linear combination of the input representation and represents how much attention will be focused on each utterance according to the dialogue act sequence. We can model the structural dependencies distribution over the latent $z$ with a linear chain CRF with n states:
\begin{align*}
	p(z_1,...,z_n|u,y)=softmax(\sum_{i=1}^{n} \theta_{i}(z_j)) \tag{8}
\end{align*}
where $\theta_{k,l}$ is the pairwise potential for $z_i=k$ and $z_{j}=l$. Notice that the utterance $u$ and the dialogue act sequence $y$ are both obtained from downstream learned representation. The marginal distribution $p(z_i|u)$ can be calculated efficiently in linear time via the forward-backward algorithm. These marginals further allow us to implicitly sum over the linear chain conditional random field. We refer to this type of attention layer as a $structural$ $attention$ $layer$, where we can explicitly look into the undirected graphical CRF structure to find which utterances are in the same chunk or in isolation. 

Here we define the node potentials with a unary CRF setting:
\begin{align*}
	\theta_{i}(j)=\sum_{j=1}^{n}u_i^T(y_j) \tag{9}
\end{align*}
where for each utterance we summarize the possible dialogue act to perform sequential reasoning. Given the potential, we compute the structural marginals $p(z_1,...,z_n|u,y)$ using the forward-backward algorithm, which is then used to compute the final probability of predicting the sequence of dialogue acts as:
\begin{align*}
	p(y_1,y_2,...,y_n, u_1,u_2,...,u_n;\theta)\\
	=\frac{\prod_{j=1}^n\sum_{j=1}^{n}u_i^T(y_j)}{\sum_{y}\prod_{j=1}^n\sum_{j=1}^{n}u_i^T(y_j)} \tag{10}
\end{align*}

\subsection{End-to-End Training}
We adopt the maximum likelihood training estimation to learn the CRF-attentive structured parameters. Given the training set $D$ with $(U,Y)$ conversation pairs, the log likelihood can be written as:
\begin{align*}
	L=\sum_{i=1}^{N}\log p(Y_i|U_i,\Theta) \tag{11}
\end{align*}
where we denote the $\Theta$ as the set of parameters within neural networks from hierarchical layers: word embedding layer, memory enhanced utterance modeling layer, CRF-attentive structured layer. We define the objective function in training process:
\begin{equation}
\underset{min}{\Theta}L(\Theta)=L(\Theta)+\lambda \left \| \Theta \right \|_{2}^{2} \tag{12}
\end{equation}
$\lambda>0$ is a hyper-parameter to trade-off the training loss and regularization. By using SGD optimization with the diagonal variant of AdaGrad, at time step t, the parameter $\Theta$ is updated as follows:
\begin{equation}
\Theta_t=\Theta_{t-1} - \frac{\rho }{\sqrt{\sum_{i=1}^{t}g_i^2}}g_t \tag{13}
\end{equation}
where $\rho$ is the initial learning rate and $g_t$ is the sub-gradient at time t.

Notice that one of our contributions is to apply CRF structural attention as the final layer of deep models. The whole model can be trained in an end-to-end manner. Here we consider the standard Viterbi algorithm for computing the distribution $p(z_1,...,z_n|u,y;\theta)$. The main procedure is summarized in Algorithm 1.

For testing, we adopt Viterbi algorithm to obtain the optimal sequence by using dynamic programming techniques. The testing procedure can be written as:
\begin{align*}
	y'=\arg \underset{y\in Y}{max} p(y|U,\Theta) \tag{14}
\end{align*} 

\begin{algorithm}[t]
	\caption{\textbf{Viterbi algorithm for CRF-ASN}}
	\begin{algorithmic}[1]
		\REQUIRE ~~\\ 
	    The observation space $O=\left\lbrace o_1,o_2,...,o_N\right\rbrace $ \\
	    The state space $S=\left\lbrace s_1,s_2,...,s_K\right\rbrace $ \\
		The observation sequence $Y=(y_1,y_2,...,y_T)$ \\
	    The probabilities $\prod=(\pi_1,\pi_2,...,\pi_K)$
	    \ENSURE ~~\\
	    The most likely hidden state sequence $X=(x_1,x_2,...,x_N)$ \\
	    ~~\\
	    \STATE Construct transition matrix $A$, each element stores the transition probability of transiting from state $s_i$ to state $s_j$
	    \STATE Construct emission matrix $B$, each element stores the probability of observing $o_j$ from state $s_i$
 		\FOR{each state $i \in \left\lbrace 1,2,...,K\right\rbrace $} 
 		       \STATE $T_1[i,1]\leftarrow\pi_i B_{iy_1}$
               \STATE $T_2[I,1]\leftarrow 0$ 		
		\ENDFOR
		\FOR{each observation $i=2,3,...,T$}
	           \FOR{each state $j\in \left\lbrace 1,2,...,K\right\rbrace $}
	                   \STATE $T_1[j,i] \leftarrow \max(T_1[k,i-1]\cdot A_{kj}\cdot B{jy_i})$    
                       \STATE $T_2[j,i] \leftarrow argmax(T_1[k,i-1]\cdot A_{kj}\cdot B_{jy_i})$                                                       	           
	           \ENDFOR
		\ENDFOR
		\STATE $z_T \leftarrow argmax(T_1[k,T])$
		\STATE $X_T \leftarrow s_{Z_T} $
		\FOR{$I \leftarrow T,T-1,...,2$}
		      \STATE $Z_{i-1} \leftarrow T_2[z_i,i]$ 
		      \STATE $X_{i-1} \leftarrow S_{Z_{i-1}}$
		\ENDFOR
		\RETURN X
	\end{algorithmic}
\end{algorithm}

\begin{table*}
	\begin{tabular}{ccl}
		\toprule
		Tag    &Example &proportion       \\
		\midrule
		STATEMENT &"I am working on my projects trying to graduate." &36\% \\
		BACKCHANNEL/ACKNOWLEDGE &"Uh-huh."   "Yeah."  "All right." "Ok..."  "Well..." &19\% \\
		OPINION &"I think it's great." / "I don't believe it can work."&13\% \\
		ABANDONED/UNINTERPRETABLE &"So, -"  "Are yo-"  "Maybe-"&6\% \\
		AGREEMENT/ACCEPT &"That's exactly it." "I can't agree more."&5\% \\
		\bottomrule
	\end{tabular}
	\caption{Top five percentages of utterance type in the SWDA corpus}
\end{table*}
\begin{table}
	\begin{tabular}{ccl}
		\toprule
		Tag    &Example &proportion \\
		\midrule
		Disruption &"yeah | he == ." "yeah | it's uh == " &14.73\% \\
		BackChannel & "okay," "right," "oh," "yes,"
		"yeah,"  &10.20\% \\
		FloorGrabber &"let's see,"
		"well," "I mean," "but.."   &12.40\% \\
		Question &Y/N, WH, Or &7.20\% \\
		Statement&"Beijing is the capital of China" &55.46\% \\
		\bottomrule
	\end{tabular}
	\caption{Top five percentages of utterance type in the MRDA corpus}
\end{table}

\section{Experiments}
In this section, we conduct several experiments on two public DA datasets SwDA and MRDA, and show the effectiveness of our approach \textbf{CRF-ASN} for dialogue act recognition.

\subsection{Data Preparation}
We evaluate the performance of our method on two benchmark DA datasets: Switchboard Dialogue Act Corpus (SwDA) and The ICSI Meeting Recorder Dialogue Act Corpus (MRDA). These two datasets have been widely used to conduct the dialogue act recognition or the dialogue act classification tasks by several prior studies. 

\begin{itemize}
\item \textbf{SwDA}: Switchboard Dialogue Act Corpus is a large hand-labeled dataset of 1155 conversations from the Switchboard corpus of spontaneous human-to-human
telephone speech. Each conversation involved two randomly selected strangers
who had been charged with talking informally about one of several, self-selected general interest topics. For each utterance, together with a variety of automatic and semiautomatic tools, the tag set distinguishes 42 mutually
exclusive utterance types via DAMSL taxonomy. The top five frequent DA types include STATEMENT, BACKCHANNEL / ACKNOWLEDGE, OPINION, ABANDONED / UNINTERPRETABLE, AGREEMENT / ACCEPT. We list the top five percentages of utterance type in the overall corpus in table2.

\item \textbf{MRDA}: The ICSI Meeting Recorder Dialogue Act Corpus consists of hand-annotated dialog act, adjacency pair, and hotspot labels for the 75 meetings in the ICSI meeting corpus. The MRDA scheme provides several class-maps and corresponding scripts for grouping related tags together into smaller number of DAs. In this work we use the most widely used class-map that groups all tags into 5 DAs, i.e., Disruption (D) indicates the current Dialogue Act is interrupted. BackChannel (B) are utterances which are not made directly by a speaker as a response and do not function in a way that elicits a response either. FloorGrabber (F) are dialogue acts for grabbing or maintaining the floor. Question (Q) is for eliciting listener feedback. And finally, unless an utterance is completely indecipherable or else can be further described by a general tag, then its default status is Statement (S). We respectively list the percentage of the five general dialogue acts in table 3.
\end{itemize} 

From the table 2 and table 3, we can see the datasets are highly imbalanced in terms of label distributions. The dialogue act type STATEMENT occupies the largest proportion in both two datasets. Following the second place is the BACKCHANNEL act type which somewhat reflect the speaker's speech style.  

\begin{table}[t]
	\small
	\begin{tabular}{|p{0.9cm}|p{0.3cm}|p{0.5cm}|l|l|l|}
		\hline
		Dataset & $|C|$ & $|V|$ & Training 	& Validation & Testing	\\\hline
		SwDA	& 42	& 19K	& 1003(173K)& 112(22K)	 & 19(4K) 	\\
		MRDA	& 5		& 10K	& 51(76K)	& 11(15K)	 & 11(15K) 	\\
		\hline
	\end{tabular}
	\caption{$|C|$ is the number of Dialogue Act classes, $|V|$ is the vocabulary size. Training, Validation and Testing indicate the number of conversations (number of utterances) in the respective splits.}
	\label{table:datastats}
\end{table}

We present the detailed data preparation procedure for obtaining the clear dataset. For two datasets, we performed pre-processing steps in order to filter out the noise and some informal nature of utterances. We first strip the exclamations and commas, and then we convert the characters into lower-case. Notice that for SwDA, we only get the training and testing datasets. In order to smooth the training step and tune the parameters, we depart the original training dataset into two parts, one for training and the other small part used to be the validation set. We list the detailed statistics of the two datasets in table 4. 

\subsection{Evaluation Criteria}
We mainly evaluate the performance of our proposed CRF-ASN method based on the  widely-used evaluation criteria for dialogue act recognition, Accuracy. The Accuracy is the normalized criteria of accessing the quality of the predicted dialogue acts based on the testing utterance set $u_t$. Given the testing conversation $C=[u_1,u_2,...,u_n]$ with its ground-truth dialogue acts $Y=[y_1,y_2,...,y_n]$, we denote the predicted dialogue acts from our CRF-ASN method by $a$. We now introduce the evaluation criteria below.
\begin{align*}
	Accuracy=\frac{1}{|u|}\sum_{u_i \in u}(1-\prod_{i=1}^{n}1[a_i\neq y_i]) \tag{15}
\end{align*}

\subsection{Implemental Details}
We preprocess each utterance using the library of nltk~\cite{loper2002nltk} and exploit the popular pretrained word embedding Glove with 100 dimensional vectors~\cite{pennington2014glove}. The size of char-level embedding is also set as 100-dimensional and is obtained by CNN filters under the instruction of Kim~\cite{kim2016character}. The Gated Recurrent Unit~\cite{cho2014properties} which is variant from LSTM~\cite{hochreiter1997long} is employed throughout our model. We adopt the AdaDelta~\cite{zeileradadelta} optimizer for training with an initial learning rate of 0.005. We also apply  dropout~\cite{srivastava2014dropout}between layers with a dropout rate of 0.2. For the memory network enhanced reasoning, we set the number of hops as 1 to preliminary learn the contextual dependencies among utterances. We do not set too many hops as increasing the number of GRU layers reduced the accuracy of the model. Early stopping is also used on the validation set with a patience of $5$ epochs. Conversations with the same number of utterances were grouped together into mini-batches, and each utterance in a mini-batch was padded to the maximum length for that batch. The maximum batch-size allowed was $48$. During training, we set the moving averages of all weights as the exponential decay rate of 0.999~\cite{lucas1990exponentially}. The whole training process takes approximately 14 hours on a single 1080Ti GPU. All the hyper-parameters were selected by tuning one hyper-parameter at a time while keeping the others fixed. 


\subsection{Performance Comparisons}
We compare our propose method with other several state-of-the-art methods for the problem of dialogue act recognition as follows:
\begin{itemize}
	\item \textbf{Bi-LSTM-CRF}~\cite{Kumar2017Dialogue} method builds a hierarchical bidirectional LSTM as a base unit and the conditional random field as the top layer to do the dialogue act recognition task.
	\item \textbf{DRLM-Conditional}~\cite{ji2016latent} method combines postive aspects of neural network architectures with probabilistic graphical models. The model combines a recurrent neural network language model with a latent variable model over shallow discourse structure.
	\item \textbf{LSTM-Softmax}~\cite{Khanpour2016Dialogue} method applies a deep LSTM structure to classify dialogue acts via softmax operation. The authors claim that the word embeddings, dropout, weight decay and number of LSTM layers all have large effect on the final performance. 
	\item \textbf{RCNN}~\cite{blunsom2013recurrent} method composes both sentence model and discourse model to extend beyond the single sentence. The authors propose hierarchical CNN on sentence model and RNN on the contextual discourses.
	\item \textbf{CNN}~\cite{lee2016sequential} method incorporates the preceding short texts to classify dialogue act. The authors demonstrate that adding sequential information improves the quality of the predictions.
	\item \textbf{HMM}~\cite{stolcke2006dialogue} method treats the discourse structure of a conversation as a hidden Markov model and the individual dialogue acts as observations emanating from the model states.
	\item \textbf{CRF} Simple baseline which applies the text encoding and CRF-based structure prediction on the DAR problem.
	\item \textbf{SVM} Simple baseline which applies the text encoding and multi-classification algorithm on the DAR problem.
	
\end{itemize}
Among them, The former five approaches eg. Bi-LSTM-CRF, DRLM-Conditional, LSTM-Softmax, RCNN, CNN all adopt the deep neural network model in order to better capture the utterances semantic representations. The latter three methods (HMM, CRF, SVM) just employ the simple feature selection on the text processing. About half of the baselines including Bi-LSTM-CRF, DRLM-Conditional, HMM, CRF consider the graphical structured prediction while the others eg. RCNN, CNN, LSTM-Softmax, SVM just adopt the traditional multi-classification algorithms.

\begin{table}[t]
	\center
	\small
	\begin{tabular}{| c | c |}
		\hline
		\textbf{Model} 						& \textbf{Accuracy(\%)} 	\\
		\hline
		Human annotator &84.0 \\
		\hline
		\textbf{Ours (CRF-ASN)} &\textbf{81.3}\\
		Bi-LSTM-CRF&79.2  \\
		DRLM-Conditional &77.0 \\
		LSTM-Softmax &75.8 \\
		RCNN &73.9\\
		CNN & 73.1\\
		CRF &71.7\\
		HMM &71.0 \\
		SVM &70.6 \\
		\hline
	\end{tabular}
	\caption{Comparing Accuracy of our method (CRF-ASN) with other methods in the literature on SwDA dataset. }
\end{table}
\begin{table}[t]
	\center
	\small
	\begin{tabular}{| c | c |}
		\hline
		\textbf{Model} & \textbf{Accuracy(\%)} \\
		\hline
		\textbf{Ours (CRF-ASN)} &91.7 \\
		Bi-LSTM-CRF & 90.9 \\
		LSTM-Softmax & 86.8\\
		CNN & 84.6 \\
		CRF & 83.9 \\
		SVM & 82.0 \\
		\hline
	\end{tabular}
	\caption{Comparing Accuracy of our method (CRF-ASN) with other methods in the literature on the MRDA dataset.}
\end{table}
Table 5 and Table 6 respectively show the experimental Accuracy results of the methods on the SwDA and MRDA datasets. The hyper-parameters and parameters which achieve the best performance on the validation set are chosen to conduct the testing evaluation. The experiments reveal some interesting points:
\begin{itemize}
	\item The results show that our proposed model CRF-ASN obviously outperforms the state-of-the-art baselines on both SwDA and MRDA datasets. Numerically, Our model improves the DAR accuracy over Bi-LSTM-CRF by 2.1\% and 0.8\% on SwDA and MRDA respectively. It is remarkable that our CRF-ASN method is nearly close to the human annotators' performance on SwDA, which is very convincing to prove the superiority of our model.
	\item The deep neural networks outperform the other feature-based models. We can see the last three non-deep models obtain worse performance than the top five deep-based methods. This suggests that the performance of dialogue act recognition can be improved significantly with discriminative deep neural networks, either in convolutional neural network or the recurrent neural network.
	\item Apart from deep learning tactics, the problem formulations are also critical to the DAR problem. We see structured prediction approaches eg. CRF-ASN, Bi-LSTM-CRF obtain better results than multi-classification eg. LSTM-Softmax. What's more, under the same text encoding situation, the CRF-based model achieves much better results than the SVM-based method. Which can fully prove the superiority of the structured prediction formulation. We also notice that CRF is better than HMM when adopted to the DAR task.
	\item The major differences between our proposed model CRF-ASN and the strong baseline BI-LSTM-CRF lie in two aspects: First we adopt a more fine grained manner to encode the utterances and utilize the memory enhanced mechanism to compute the contextual dependencies. Second we employ an adapted structured attention network on the CRF layer, rather than directly apply the original CRF on the utterances. These two modifications are essential and improve the performance significantly.
\end{itemize}
\begin{table}[t]
	\center
	\small
	\begin{tabular}{| c | c |}
		\hline
		\textbf{Model} & \textbf{Accuracy(\%)} \\
		\hline
		\textbf{Full CRF-ASN} &\textbf{81.3} \\
		\hline
		Simple CRF & 79.5 \\
		Simple SVM &77.8 \\
		Simple Word Embedding &78.7\\
		Simple Context state &79.1\\
		Simple Memory Network &78.3\\
		Simple Utterance Embedding &79.0\\
		\hline
	\end{tabular}
	\caption{Component ablations on SwDA dataset}
\end{table}
\subsection{Ablation Results}
We respectively evaluate the individual contribution of the proposed module in our model. We conduct thorough ablation experiments on the SwDA dataset, which are recorded on the table 7. To make it fair, we only modify one module at a time and fix the other components to be in the same settings.

\begin{itemize}
	\item We replace the proposed structured CRF-attention layer to simple CRF, the results show structured CRF-attention layer results in major improvement in the accuracy, approximately over 2.1\% absolute points. We further replace the structure prediction formulation to multi-classification on SVM, the results drop dramatically, which illustrate the benefit of considering structural dependencies among utterances.
	\item We replace the fine-grained word $E_w, E_c, POS, NER$ to the simple Glove vector. The results suggest that fine grained word embedding is useful to represent a text. We also adapt the context state $h_t$ to only care its neighbor utterances. The result is not satisfying, which conveys us that the basic text understanding is critical in the semantic representations.
	\item We replace the memory network to directly apply CRF layer to the utterance layer. We also conduct a comparing experiment which plus the original utterance to memory enhanced output. The two results show the designed hierarchical memory-enhanced components are helpful in the utterance understanding and modeling the contextual influence. 
\end{itemize}

\subsection{Visualization}
\begin{figure}[t]
	\centering
	\includegraphics[width=8cm]{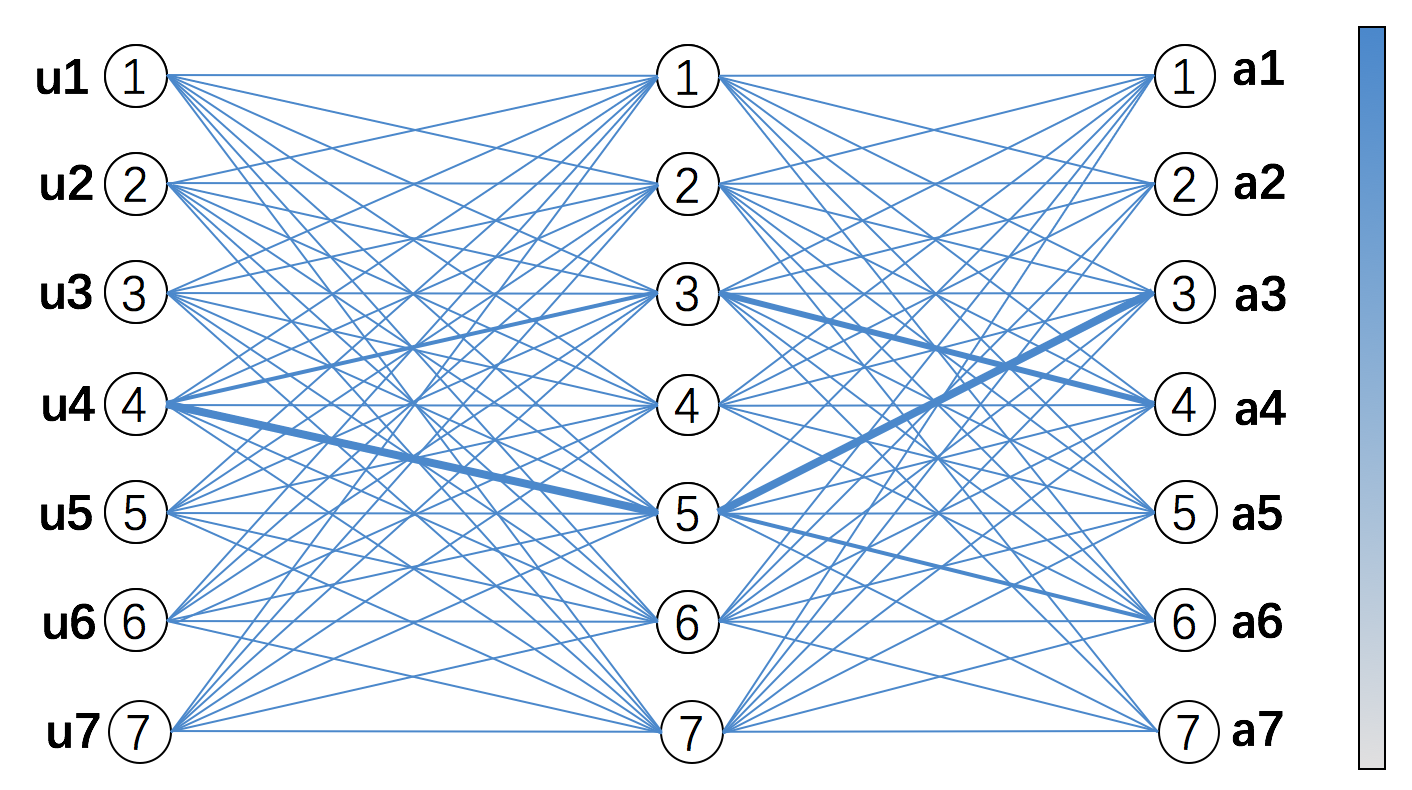}
	\caption{Visualization of the structured attention distribution over conditional random field sequence. For example, the edges represent the marginal probabilities $p(z|u,\Theta)$, and the nodes represent the utterances and corresponding dialogue acts. In this figure we can see For utterance u4, dialogue a3 is the most suitable predicting label as the edge ($4\rightarrow5\rightarrow3$) is the most attentive path.}
\end{figure}

\begin{figure}[t]
	\centering
	\includegraphics[width=10cm]{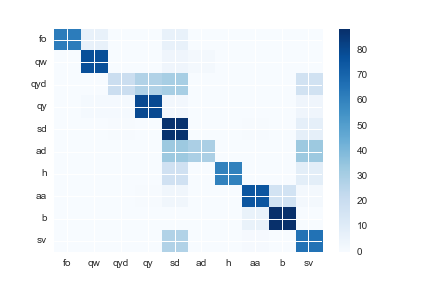}
	\caption{Confusion heatmap of CRF-ASN model for the SwDA dataset. There are totally 10 DA labels, where the row denotes the true label and the column denotes the predicted label.}
\end{figure}

In Figure 3, we visualize of the output edge marginals produced by the CRF-ASN model for a conversation. In this instance, the actual dialogue act recognition procedure is displayed as $4\rightarrow5\rightarrow3$. In the testing step, the model is uncertain and select the most attentive path to maximize the true dialogue act recognition. Here we can see from the marginal edges the path $4\rightarrow5\rightarrow3$ occupies more attentive weights than the path  $4\rightarrow3\rightarrow4$ in predicting the dialogue act label. Thus we ultimately select the right way to recognize the dialogue act.

Figure 4 shows the confusion heatmap of our proposed CRF-ASN model for the SwDA dataset. Each element in the heatmap denotes the rate that the predicted label is the same to the true label. We can see from the diagonal, the <sd,sd> <b,b> pairs achieve the most satisfying matching score while <qyd, qyd> is much worse than other pairs. This can be explained that the sd (statement) and b(acknowledge) have clearly self-identification while qyd(Declarative Yes-No-Question) is more easier to be mistakenly recognized. We can see that <qyd,qy> which represents (Declarative Yes-No-Questio,Yes-No-Question) is indeed hard to recognize since their dialogue type are too similar with each other. For another reason, we notice that due to the bias of the ground truth, there are some cases that we predict the dialogue act correctly while the ground truth is wrong. To some reason, classifying so many fine-grained dialogue act labels is not easy for human annotators, besides the human-subjectivity occupies an important role in recognizing the dialogue act.

\section{Related Work}
In this section, we briefly review some related work on dialogue act recognition and attention network. 

\subsection{Dialogue Act Recognition}
The main task of dialogue act recognition is to assign an act label to each utterance in a conversation, which can be defined as a supervised problem due to the properties that each utterance has a corresponding act label. Most of the existing work for the problem of dialogue act recognition can be categorized as following two groups.

\textbf{Regarding the DAR as a multi-classification problem}.  Reithinger et al.~\cite{reithinger1997dialogue} present deal with the dialogue act classification using a statistically based language model. Webb et al.~\cite{webb2005dialogue} apply diverse intra-utterance features involving word n-gram cue phrases to understand the utterance and do the classification. Geertzen et al.~\cite{geertzen2007multidimensional} propose a multidimensional approach to distinguish and annotate units in dialogue act segmentation and classification. Grau et al.~\cite{grau2004dialogue} focus on the dialogue act classification using a Bayesian approach. Serafin et al.~\cite{serafin2003latent} employ Latent Semantic Analysis (LSA) proper and augmented method to work for dialogue act classification. Chen et al.~\cite{chen2013empirical} had an empirical investigation of sparse log-linear models for improved dialogue act classification. Milajevs et al.~\cite{milajevs2014investigating} investigate a series of compositional distributional semantic models to dialogue act classification. 

\textbf{Regarding the DAR as a sequence labeling problem}. 
Stolcke et al.~\cite{stolcke2006dialogue} treat the discourse structure of a conversation as a hidden Markov model and the individual dialogue acts as observations emanating from the model states. Tavafi et al.~\cite{tavafidialogue} study the effectiveness of supervised learning algorithms SVM-HMM for DA modeling across a comprehensive set of conversations. Similar to the SVM-HMM, Surendran et al.~\cite{surendran2006dialog} also use a combination of linear support vector machines and hidden markov models for dialog act tagging in the HCRC MapTask corpus. Lendvai et al.~\cite{lendvai2007token} explore two sequence learners with a memory-based tagger and conditional random fields into turn-internal DA chunks. Boyer et al.~\cite{boyer2009discovering} also applied HMM to discover internal dialogue strategies inherent in the structure of the sequenced dialogue acts.  Galley et al.~\cite{galley2006skip} use skip-chain conditional random field to model non-local pragmatic dependencies between paired utterances. Zimmermann et al.~\cite{zimmermann2009joint} investigate the use of conditional random fields for
joint segmentation and classification of dialog acts exploiting both word and prosodic features.   

Recently, approaches based on deep learning methods improved many state-of-the-art techniques in NLP including DAR accuracy on open-domain conversations~\cite{Kalchbrenner2013Recurrent}~\cite{zhou2015combining}~\cite{Ji2016ALV}~\cite{Kumar2017DialogueAS}~\cite{lee2016sequential}. Kalchbrenner et al.~\cite{Kalchbrenner2013Recurrent} used a mixture of CNN and RNN. CNNs were used to extract local features from each utterance and RNNs were used to create a general view of the whole dialogue. Khanpour et al.~\cite{Khanpour2016Dialogue} design a deep neural network model that benefits from pre-trained word embeddings combined with a variation of the RNN structure for the DA classification task. Ji et al.~\cite{Ji2016ALV} also investigated the performance of using standard RNN and CNN on DA classification and got the cutting edge results on the MRDA corpus using CNN. Lee et al.~\cite{lee2016sequential} proposes a model based on CNNs and RNNs that incorporates preceding short texts as context to classify current DAs. Zhou et al.~\cite{zhou2015combining} combine heterogeneous information with conditional random fields for Chinese dialogue act recognition. Kumar et al.~\cite{Kumar2017DialogueAS} build a hierarchical encoder with CRF to learn multiple levels of utterance and act dependencies. 

Unlike the previous studies, we formulate the problem from the viewpoint of integrating contextual dependencies in both utterance level and the act label level. We not only consider the fine grained multi-level semantic representations, but also integrate the structured attention network to further capture the structure designpendencies in the CRF layer.
\subsection{Attention Network}
Attention mechanism has become an essential component in text understanding in recent years. Since the first work proposed by Bahdanau et al.~\cite{Bahdanau2014Neural} that  adopt the attention mechanism in neural machine translation, attention mechanism based neural networks have become a major trend in diverse text researching field, such as in machine comprehension~\cite{hermann2015teaching}~\cite{yinattention}~\cite{kadlectext}~\cite{dhingragated}, machine translation~\cite{luongeffective}~\cite{firat2016multi}, abstract summarization~\cite{rushneural}~\cite{allamanis2016convolutional}, text classification~\cite{wang2016relation}~\cite{zhou2016attention}~\cite{yang2016hierarchical} and so on. The principle of attention mechanism is to select the most pertinent piece of information, rather than using all available information, a large part of it being irrelevant to compute the neural response.

In our work, we propose the CRF-attentive structured network in order to encode the internal utterance inference with dialogue acts. The structured attention is a more general attention mechanism which take account of the graphical dependencies and allow for extending attention beyond the standard soft-selection approach. The most similar work to our model is proposed by Kim et al.~\cite{kimstructured}. Kim et al. also experiment with two different classes of structured attention networks: subsequence selection and syntactic selection. However, the objectives of these two networks aims to segment the structure dependencies, which are quite different from our DAR task. In DAR task we care more on the dialogue act influences on the overall conversation structure, thus the former structured attention may not be suitable for our problem.

\section{Conclusion}
In this paper, we formulate the problem of dialogue act recognition from the viewpoint of capturing hierarchical rich utterance representations and generalize richer CRF attentive graphical structural dependencies without abandoning end-to-end training. We propose the CRF-Attentive Structured Network (CRF-ASN) for the problem. We implement the model in two steps. We first encode the rich semantic representation on the utterance level by incorporating hierarchical granularity and memory enhanced inference mechanism. The learned utterance representation can capture long term dependencies across the conversation. We next adopt the internal structured attention network to compute the dialogue act influence and specify structural dependencies in a soft manner. This approach enable the soft-selection attention on the structural CRF dependencies and take account of the contextual influence on the nearing utterances. We demonstrate the efficacy of our method using the well-known public datasets SwDA and MRDA. The extensive experiments demonstrate that our model can achieve better performance than several state-of-the-art solutions to the problem.   
\bibliographystyle{ACM-Reference-Format}
\bibliography{sample-bibliography} 
\end{document}